\documentclass[letterpaper, 10 pt, conference]{ieeeconf}  

\IEEEoverridecommandlockouts                              

\usepackage{graphics} 
\usepackage{epsfig} 
\usepackage{mathptmx} 
\usepackage{times} 
\usepackage{amsmath} 
\usepackage{amssymb}  
\usepackage{xcolor,graphicx}
\usepackage{lipsum}
\usepackage{booktabs}
\usepackage{siunitx}
\usepackage{multirow} 
\usepackage[noadjust]{cite}

\title{\LARGE \bf
Assist-as-needed Control for FES in Foot Drop Management
}

\author{Andreas Christou$^{1}$, Elliot Lister$^{1}$, Georgia Andreopoulou$^{2}$, Don Mahad$^{2}$, and Sethu Vijayakumar$^{1}$
\thanks{This research was supported in part by the Engineering and Physical Sciences Research Council (EPSRC, grant reference EP/Y030834/1) as part of the Centre for Doctoral Training Dependable and Deployable AI for Robotics at Heriot-Watt University and The University of Edinburgh, in part by the Alan Turing Institute, U.K., and in part by the Japan Science and Technology Agency (JST) Moonshot R\&D Program (Grant No. JPMJMS2239).}
\thanks{This paper has supplementary downloadable material available at http://XXXXXXXX, provided by the authors. This includes one multimedia MP4 format movie clip, which provides scenes of the experimental setup. This material is XXX MB in size.} 
\thanks{A. Christou and E. Lister are with the School of Informatics, University of Edinburgh, EH8 9BT UK, (email: andreas.christou@ed.ac.uk).}
\thanks{G. Andreopoulou and D. Mahad are with the Anne Rowling Neurology Clinic, School of Medicine, University of Edinburgh, EH16 4SB UK.}
\thanks{S. Vijayakumar is with the Alan Turing Institute, U.K., and the School of Informatics, University of Edinburgh, EH8 9BT UK.}%
}

\begin{document}

\maketitle
\thispagestyle{empty}
\pagestyle{empty}

\begin{abstract}

Foot drop is commonly managed using Functional Electrical Stimulation (FES), typically delivered via open-loop controllers with fixed stimulation intensities. While users may manually adjust the intensity through external controls, this approach risks overstimulation, leading to muscle fatigue and discomfort, or understimulation, which compromises dorsiflexion and increases fall risk. In this study, we propose a novel closed-loop FES controller that dynamically adjusts the stimulation intensity based on real-time toe clearance, providing ``assistance as needed". We evaluate this system by inducing foot drop in healthy participants and comparing the effects of the closed-loop controller with a traditional open-loop controller across various walking conditions, including different speeds and surface inclinations. Kinematic data reveal that our closed-loop controller maintains adequate toe clearance without significantly affecting the joint angles of the hips, the knees, and the ankles, and while using significantly lower stimulation intensities compared to the open-loop controller. These findings suggest that the proposed method not only matches the effectiveness of existing systems but also offers the potential for reduced muscle fatigue and improved long-term user comfort and adherence.

\end{abstract}

\section{INTRODUCTION}
Foot drop is a common symptom after neurological damage that can limit the mobility of patients and negatively impact their quality of life. Due to compromised spinal cord pathways, and/or muscle function, foot drop can manifest itself in a wide range of conditions such as stroke, multiple sclerosis, cerebral palsy and traumatic injury \cite{gil-castillo_advances_2020}. As a result, there is an increased risk of falling that can limit the confidence of walking independently and can significantly reduce gait speed. 

Functional electrical stimulation (FES) has shown great promise in preventing foot drop and facilitating independent ambulation in people experiencing foot drop \cite{Andreopoulou2024}. 
Using non-invasive electrodes and low-intensity electrical impulses, FES can externally activate the patient's inert muscles to support gait and prevent foot drop. This offers an affordable and lightweight solution to manage foot drop that can have a significant positive impact in people's daily living. 

However, the way FES is currently being used lacks personalisation and relies on the user's conscious input in order to adjust the stimulation intensity. In most cases, commercial devices can differentiate between the stance and the swing during gait, either using pressure sensors or inertial measurement units, and use this information to turn FES ON or OFF at the intensity selected by the therapist or the user \cite{gil-castillo_advances_2020}. As a result, higher or lower than needed stimulation may be provided, which may either unnecessarily fatigue the muscles or put the user at risk of falling, respectively. 

This has led to the investigation of alternative control systems that can modulate the intensity of stimulation. Proportional control has been studied in order to regulate stimulation intensity based on electromyography (EMG) \cite{Yeom2010,Xu2024} 
or the error between desired and actual movement \cite{Christou2024,anaya-reyes_omnidirectional_2020}, where the desired movement is defined based on the average ankle joint kinematics of healthy controls. Other EMG-informed controllers have also been tested where pre-defined stimulation patterns were used for the different gait phases, based on the EMG activity observed during the gait of healthy controls \cite{shaikh_bipedal_2018}. 
Iterative learning control \cite{Hosiasson2023,Seel2016} 
and repetitive control \cite{Page2020}, which refine stimulation based on previous gait cycles and previous samples, respectively, have also been studied, but the performance of these methods is highly dependent on the periodicity of gait. Model predictive control (MPC) and data-driven controllers have also been explored \cite{Singh2023,li_adaptive_2021} but these methods can be computationally expensive and may suffer from model inaccuracies or limited generalisability, respectively.  

What has not been studied extensively is how FES can be reliably used in daily life where the user may frequently change their gait speed or may walk on terrain of variable inclination. It is evident that human biomechanics can significantly change depending on these conditions \cite{reznick_lower-limb_2021,MCINTOSH20062491} and it is prudent that our FES devices take this into consideration to ensure comfort and safety. 

\begin{figure*}
    \centering
\includegraphics[width=\textwidth]{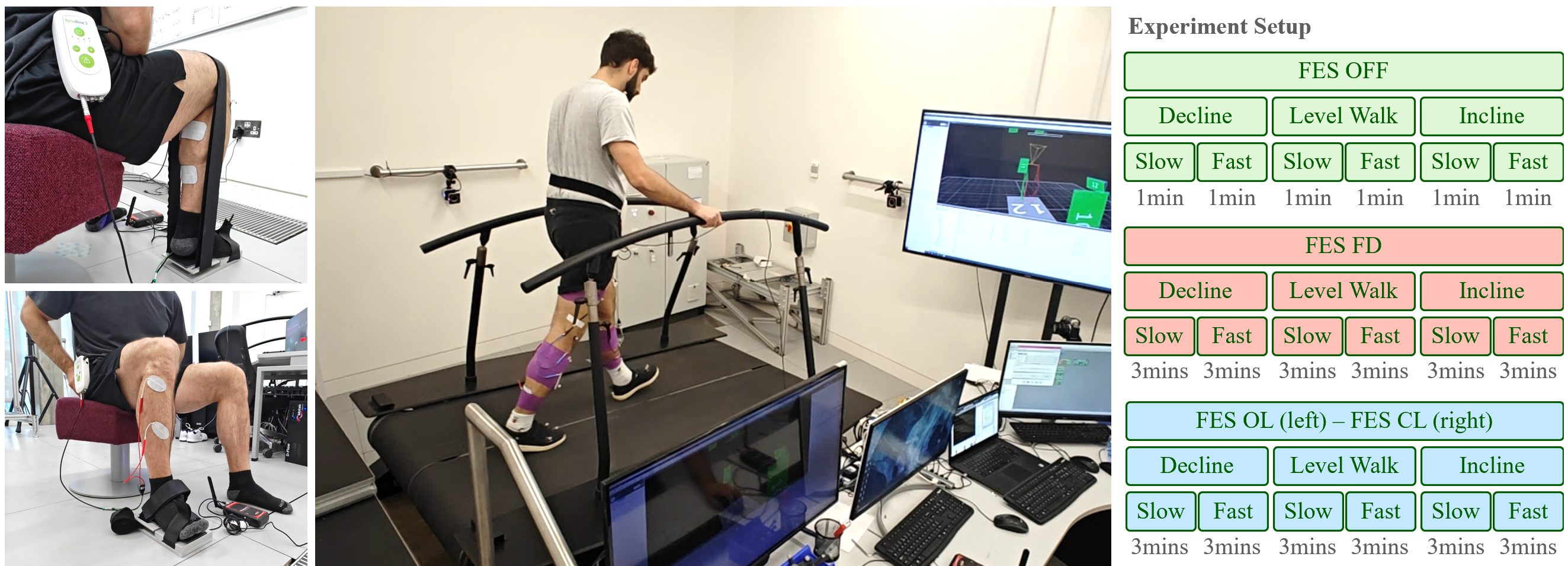}
\caption{Experimental setup including the calibration of the FES intensity for each participant for both the plantarflexor and dorsiflexor muscles (left), and marker-based motion capture using Vicon and the instrumented treadmill M-GAIT (middle), for the different conditions shown on the right.}
\label{fig:Experiment}
\end{figure*}

In this paper we present a novel method for controlling FES using toe clearance as an input in order to provide assistance as needed. We propose a closed-loop controller that adjusts the stimulation intensity in real time to ensure adequate toe clearance during gait at different speeds and slope. It is hypothesised that our proposed controller will more effectively adjust the intensity of stimulation to guarantee safety and minimise stimulation compared to an open-loop control system where the stimulation profile is kept constant. 
Our contributions include: (1) the development of a novel closed-loop FES controller, (2) the use of toe clearance as an input which is an underutilised yet more direct and informative predictor of gait safety and stability than metrics like ankle orientation, as it directly reflects the risk of stumbling, (3) the controller's validation on healthy participants, and (4) its comparison to commercial open-loop FES control (Figure \ref{fig:Experiment}). We also provide valuable kinematic data obtained from healthy controls walking at different speeds and slopes, recorded using 3D motion capture technology, and analysed using musculoskeletal modelling. 

\section{Methods}

\subsection{Subjects}
This study was carried out with the help of 6 healthy volunteers (4 male, 2 female, age = 29 $\pm$ 6, weight = 75kg $\pm$ 10kg). The experiment pipeline was approved by the University of Edinburgh, School of Informatics Ethics Committee (ID 2024/903647) and all participants provided written consent.

\subsection{Hardware}
To analyse the motion of the participants a 3D motion capture system was used, comprising of 12 Vicon cameras (UK), along with the instrumented treadmill from MOTEK Medical, M-Gait (The Netherlands). Marker data was recorded at a frequency of 100Hz and was used to adjust the intensity of stimulation at an average frequency of 98Hz. The stimulators RehaMove3 (Hasomed, Germany) and EM49 (Beurer, Germany) were used to deliver FES during the experiment and prior to the experiment, respectively. The analysis was carried out on a Windows PC using an interface between the musculoskeletal modelling software, OpenSim \cite{OpenSim}, and MATLAB. 

\subsection{Experimental Setup}
The participants visited our laboratory twice. Upon their first visit, the participants were informed about the experiment and were provided with the EM49 stimulator. Participants were asked to use the stimulator to become familiar with electrical stimulation. 
The second visit was scheduled three days after the first visit. Upon their second visit, the participants first underwent a muscle identification process. During this process, the threshold stimulation, $u_{thr}$, and maximum stimulation, $u_{max}$, were identified for each muscle using the RehaMove3 stimulator and our ankle dynamometer (Figure \ref{fig:Experiment} left) by incrementally increasing the pulse width by 50$\mu s$ at a time, while the stimulation frequency and amplitude were kept constant at 25Hz and 25mA, respectively. The duration of each signal was set to 1s, followed by a 5s break. The threshold stimulation was defined as the pulse width that generated a measurable muscle force, and the maximum stimulation was defined as either the point of discomfort, as indicated by the participant, or the point of muscle saturation, as observed by the generated force. 

After the identification phase, the participants were fitted with reflective markers, and were asked to step on the treadmill. First, a static posture of the participants was recorded, and then the motion of the participants was recorded under three conditions: (1) while the FES was off (FES OFF), (2) while the FES was used to stimulate the plantarflexor muscles to induce foot drop (FES FD), and (3) while the FES was used to stimulate both the plantaflexor and the dorsiflexor muscles to both induce foot drop and correct it. For the latter case, the dorsiflexors of each leg were stimulated using a different controller. An open-loop controller, generating a trapezoidal signal of fixed stimulation profile, was used for the left leg (FES OL), whereas our proposed closed-loop controller was used for the right leg (FES CL). During conditions (2) and (3), the stimulation of the targeted muscles was delivered with a probability of 0.25 to reduce the risk of encouraging compensatory mechanisms and voluntary adjustments in the participants' kinematics. In all three conditions, the participants' motion was recorded while walking on a decline (-5\textdegree), on level ground (0\textdegree) and on an incline (5\textdegree), and while walking at a slow pace (0.7m/s) and fast pace (1.2m/s). This resulted in a total of 18 recordings per participant. The duration of each recording during condition (1) was set to 1 minute, while the duration of each recording during conditions (2) and (3) was set to 3 minutes.

\subsection{Motion Capture}
\begin{figure}
    \centering
\includegraphics[width=\linewidth]{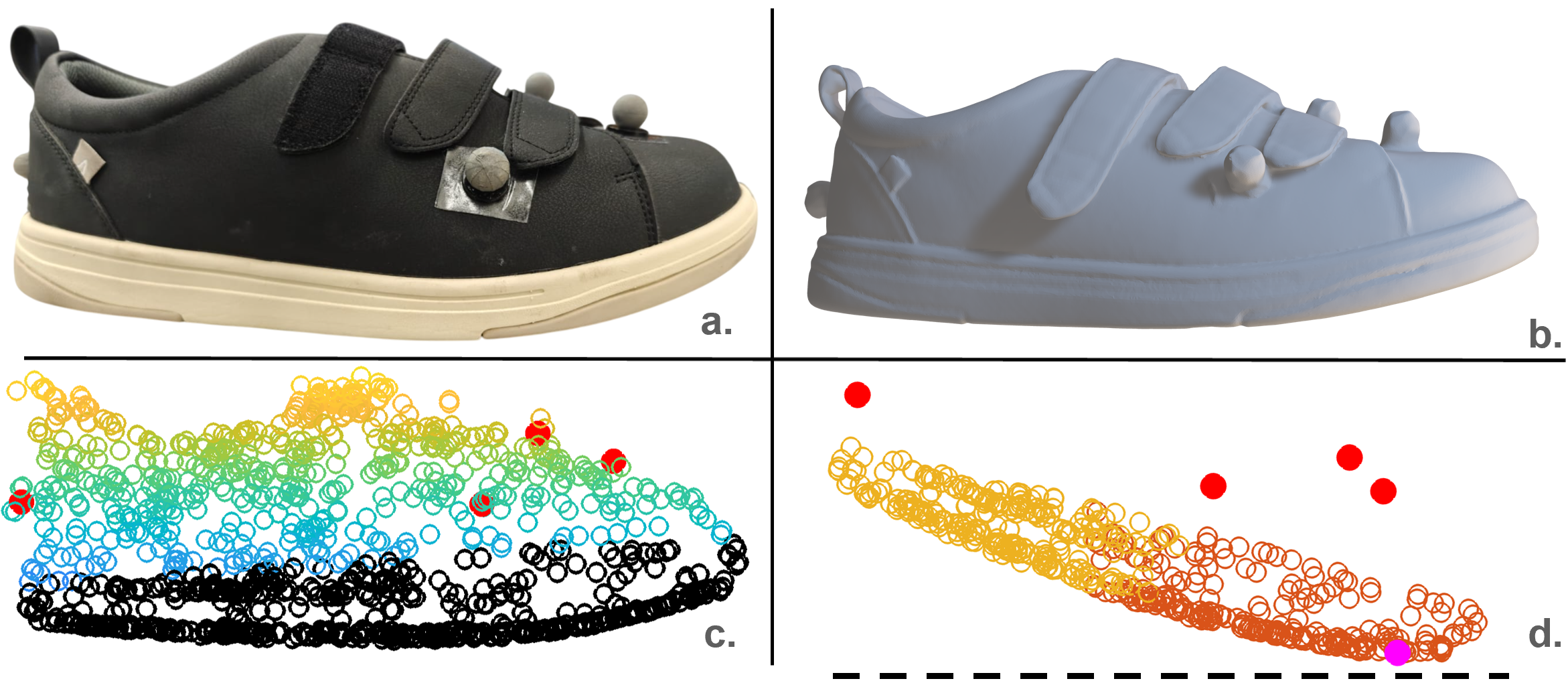}    \caption{(a) Image of the shoe with reflective markers. (b) 3D scan of the shoe including reflective markers. (c) Point cloud of the shoe where black dots indicate the base of the shoe and red dots indicate the location of the centre of the reflective markers. (d) Toe clearance estimation based on the recorded position of the reflective markers (shown in red), the anterior-half points of the base of the shoe (shown in orange) and an estimation of the ground (shown as a dashed line).}
\label{fig:Shoe3DScan}
\end{figure}

To record the motion of the participants, reflective markers were placed on the torso, the pelvis and the lower limbs (as described in \cite{Christou2025}). Anatomical markers were placed on bony landmarks, and tracking markers were included at the thighs and the shanks to improve tracking accuracy. 
The shoe markers included a marker above the first metatarsophalangeal joint (MTP1), a marker above the fifth metatarsophalangeal joint (MTP5), a marker above the second proximal phalange (PP2), and a marker at the heel (Figure \ref{fig:Shoe3DScan}).

\subsection{Toe Clearance Estimation}
Toe clearance, $c$, was measured using a 3D scan of the shoes including the reflective markers, and the recorded position of the markers obtained from Vicon. The 3D scan was obtained using the EinScan HX (Shining 3D, China) and the anterior-half of the base of the scan was used to estimate toe clearance. Toe clearance was defined as the shortest distance identified between any point on the anterior base of the shoe and the ground (Figure \ref{fig:Shoe3DScan}d). A static pose of the unloaded shoes was recorded with the treadmill at 0\textdegree\ and 10\textdegree\ during calibration to estimate the position of the ground and define zero toe clearance. 

\subsection{Stimulation controllers}
To deliver FES, gel electrodes were placed on both legs on the surface of the muscles responsible for plantarflexion and dorsiflexion (Figure \ref{fig:Experiment} left). Rectangular electrodes (87x50mm) were mostly used to target the gastrocnemius and the soleus (GS), while oval electrodes (62x40mm) were used to target the tibialis anterior (TA). 

\paragraph{Open-loop control}
Two open-loop FES controllers were used during the experiment: (1) to induce foot drop  (both legs) and (2) to correct foot drop (left leg only). To induce foot drop, the GS muscles were stimulated from mid-stance to heel strike using a trapezoidal signal, whose peak was set to just above the identified threshold intensity to generate a plantar flexion force of approximately 10-15N.

To correct for the induced foot drop, the TA of the left leg was stimulated during swing using a trapezoidal signal, whose peak was defined as:
\begin{gather}
    u^{TA_L}=u_{thr}^{TA_L}+\frac{u_{max}^{TA_L}-u_{thr}^{TA_L}}{2}
    \label{eq:TA_L_peak}
\end{gather}

It was verified through the muscle identification process, that these stimulation intensities can generate adequate force to induce foot drop and correct foot drop during the FES FD condition and FES OL condition, respectively. 

\paragraph{Closed-loop control}
The TA of the right leg was stimulated using a closed-loop controller. This controller used toe clearance as the input, and adjusted the stimulation pulse width, $u^{TA_R}$, in real time to ensure adequate clearance. A minimum toe clearance, $c_{min}$, was defined at 10mm based on the values reported in the literature \cite{Fehr2024}, and a controller threshold toe clearance, $c_{thr}$, was defined at 25mm. Based on these values the closed-loop controller was defined as:

\begin{gather}
    u=\begin{cases} k_{p}e_{min} + k_{d}\dot{e}_{min}+k_{i}\int{e_{min}}, & c<c_{min} \\
    k_{d}\text{max}(0,\dot{e}_{thr}), & c_{min}<c<c_{thr} \\
    0, & c_{thr}<c
    \end{cases}\\
    u^{TA_R}=\begin{cases}
        u^{TA_R}_{thr}, & u<u_{thr}^{TA_R} \\
        u, & u_{thr}^{TA_R}<u<u_{max}^{TA_R} \\
        u_{max}^{TA_R}, & u_{max}^{TA_R}<u
    \end{cases}
\end{gather}

where $k_{p}$, $k_{d}$, and $k_{i}$ are the controller gains (set to [25, 0.7s, 1s$^{-1}$]$\mu$s/mm respectively), and $e_{min}$ and $e_{thr}$ are the controller errors defined as $e_{min}=c_{min}-c$ and $e_{thr}=c_{thr}-c$, respectively. 
In all controllers, the rate of change of pulse width was limited to 50$\mu$s per time step, a value determined experimentally to yield smoother and more comfortable stimulation profiles.

\begin{figure*}[ht]
    \centering
    \includegraphics[width=\linewidth]{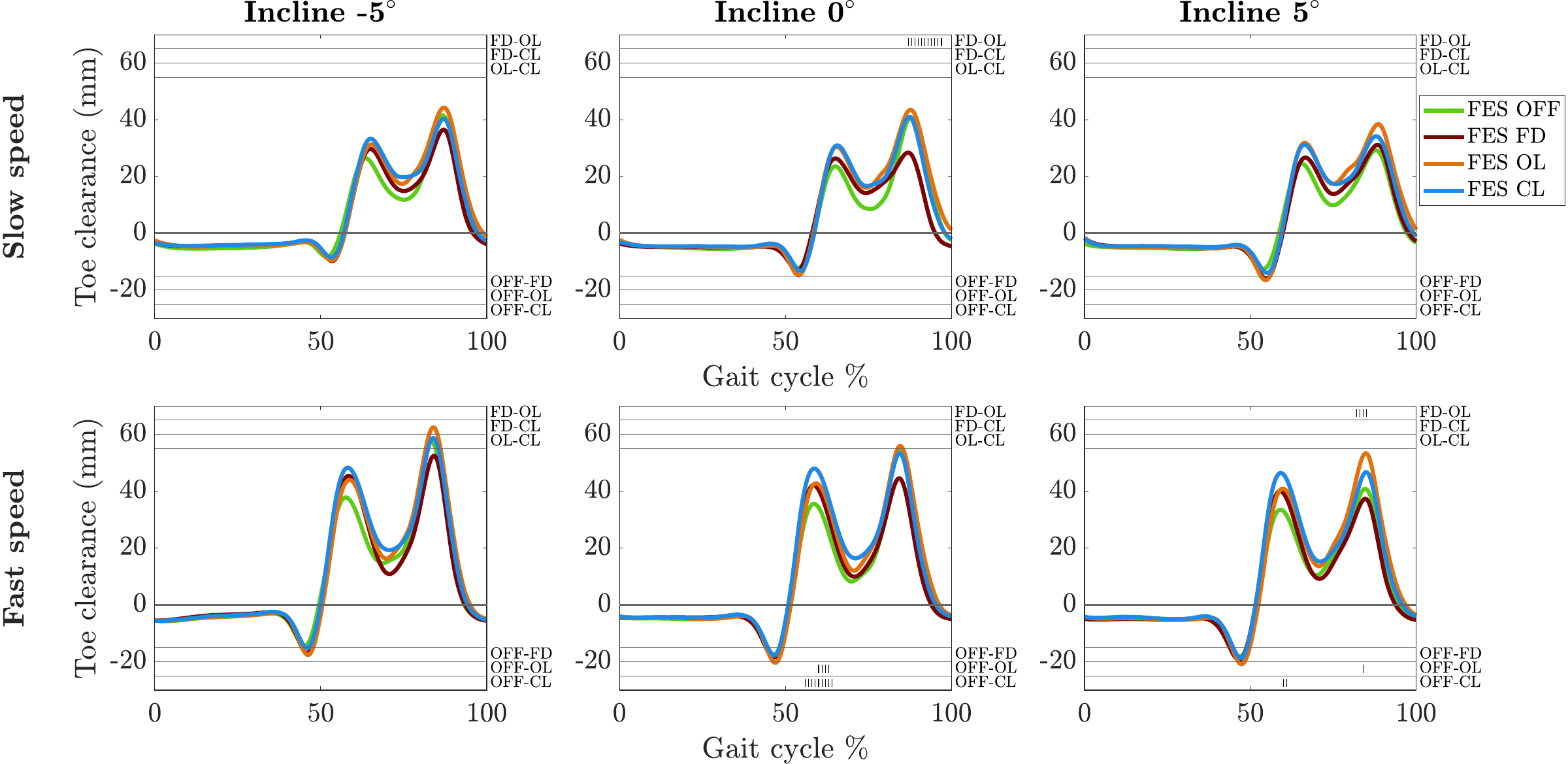} 
    \caption{Mean toe clearance ($\pm$SD) of all recorded cycles from all subjects under the different conditions including toe clearance during slow walk (top row) and fast walk (bottom row), toe clearance during decline (left column), level walking (middle column) and incline (right column), and toe clearance during FES OFF (in green), FES FD (in red), FES OL (in orange) and FES CL conditions (in blue). For clarity, the variance of the recorded toe clearance is also provided in Table \ref{tab:toe_clearance_variability}. Vertical bars ($|$) indicate points in the gait cycle where mean differences between conditions are statistically significant (p$<$0.05).}
    \label{fig:ToeClearance}
\end{figure*}

\subsection{State Estimation}
For the estimation of gait phase, marker trajectories were used. Swing and stance were defined based on the trajectory of the PP2 marker and its rate of change in the anterior-posterior direction. This was preferred over state estimation that would rely on ground-force data, which could provide higher temporal precision, but could fail to identify state transitions if participants stepped over the middle line of the force plates. The onset of mid-stance was defined based on the relative position of the PP2 marker and the markers placed on the left and right anterior superior iliac spine.

\subsection{Analysis}
To compare the effect of the different conditions on toe clearance, the recorded data was segmented into gait cycles ranging from 0-100\% and statistical tests were carried out to observe the points in the gait cycle where the participants' motion is significantly different. For this, pairwise comparisons were carried out between all four conditions using permutation tests with 10000 permutations across the different points in the gait cycle. Since toe clearance is a parameter that is affected by the whole-body kinematics, similar analysis was carried out for the motion of the hips, knees and ankles in the sagittal plane. These joint kinematics are reported with a focus on the swing phase and the mid-swing region where the minimum toe clearance occurs.





\section{RESULTS}

\begin{figure*}[ht]
    \centering
    \includegraphics[width=\linewidth]{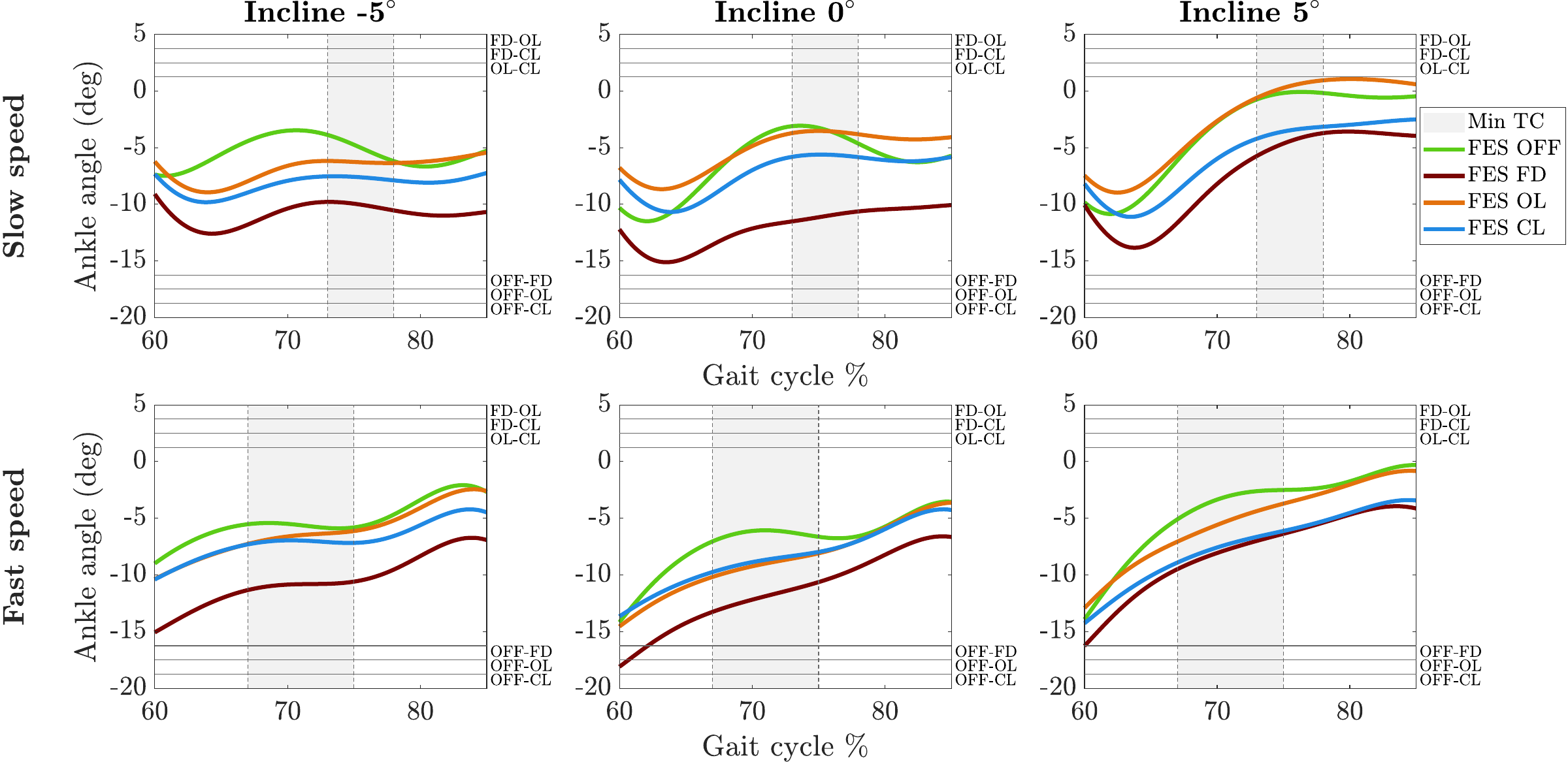}    
    \caption{Mean ankle angle during the swing phase for all subjects under the different conditions including toe clearance during slow walk (top row) and fast walk (bottom row), toe clearance during decline (left column), level walking (middle column) and incline (right column), and toe clearance during FES OFF (in green), FES FD (in red), FES OL (in orange) and FES CL conditions (in blue). The region where minimum toe clearance (Min TC) is expected, is highlighted in a grey background. Ankle angle variability is not shown for clarity. Vertical bars ($|$) indicate points in the gait cycle where mean differences between conditions are statistically significant (p$<$0.05).}
    \label{fig:Ankle Angle}
\end{figure*}

\begin{figure*}
    \centering
    \includegraphics[width=\linewidth]{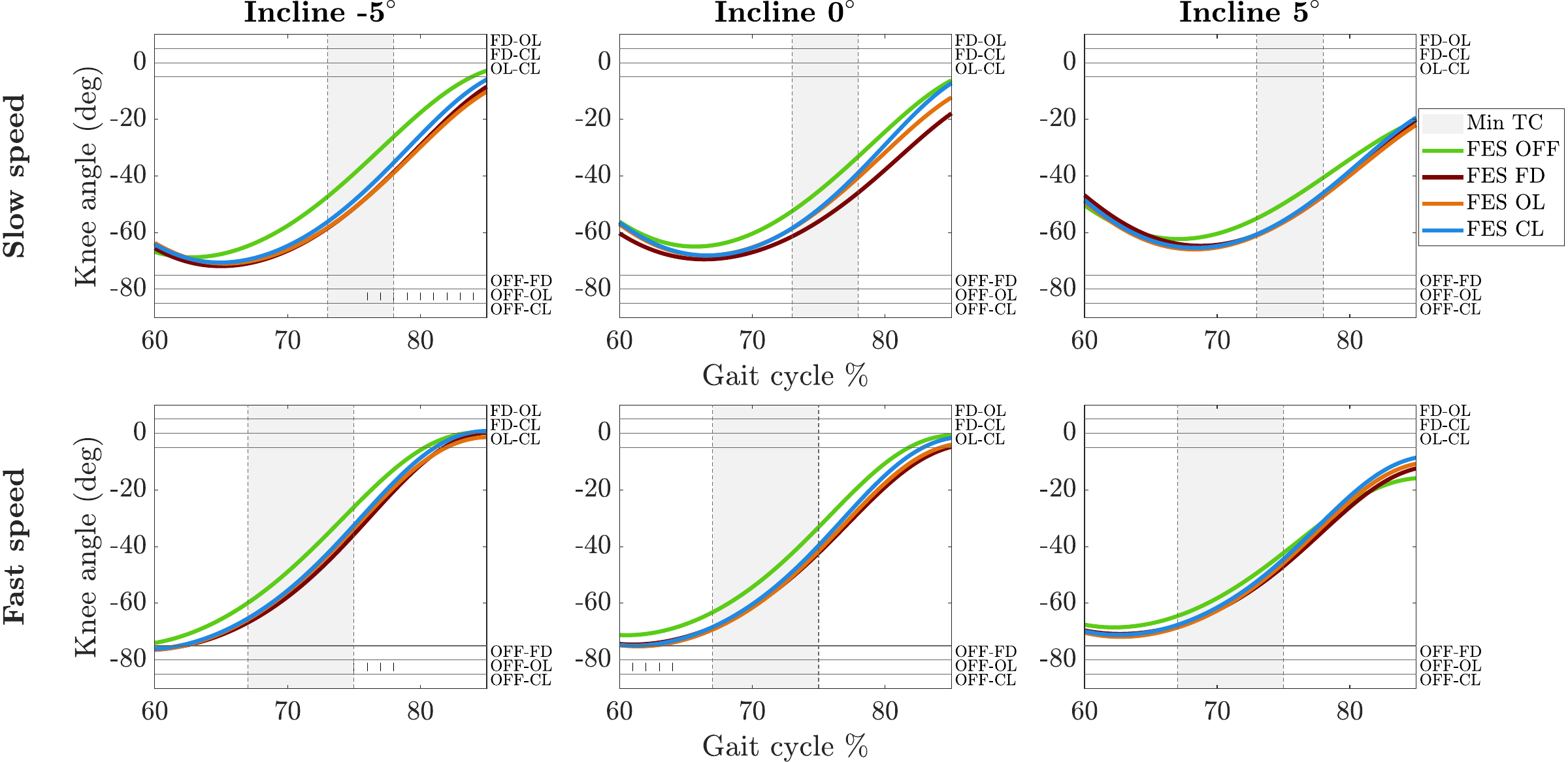}    
    \caption{Mean knee angle during the swing phase for all subjects under the different conditions including toe clearance during slow walk (top row) and fast walk (bottom row), toe clearance during decline (left column), level walking (middle column) and incline (right column), and toe clearance during FES OFF (in green), FES FD (in red), FES OL (in orange) and FES CL conditions (in blue). The region where minimum toe clearance (Min TC) is expected, is highlighted in a grey background. Knee angle variability is not shown for clarity. Vertical bars ($|$) indicate points in the gait cycle where mean differences between conditions are statistically significant (p$<$0.05).}
    \label{fig:Knee Angle}
\end{figure*}

\begin{figure*}
    \centering
    \includegraphics[width=\linewidth]{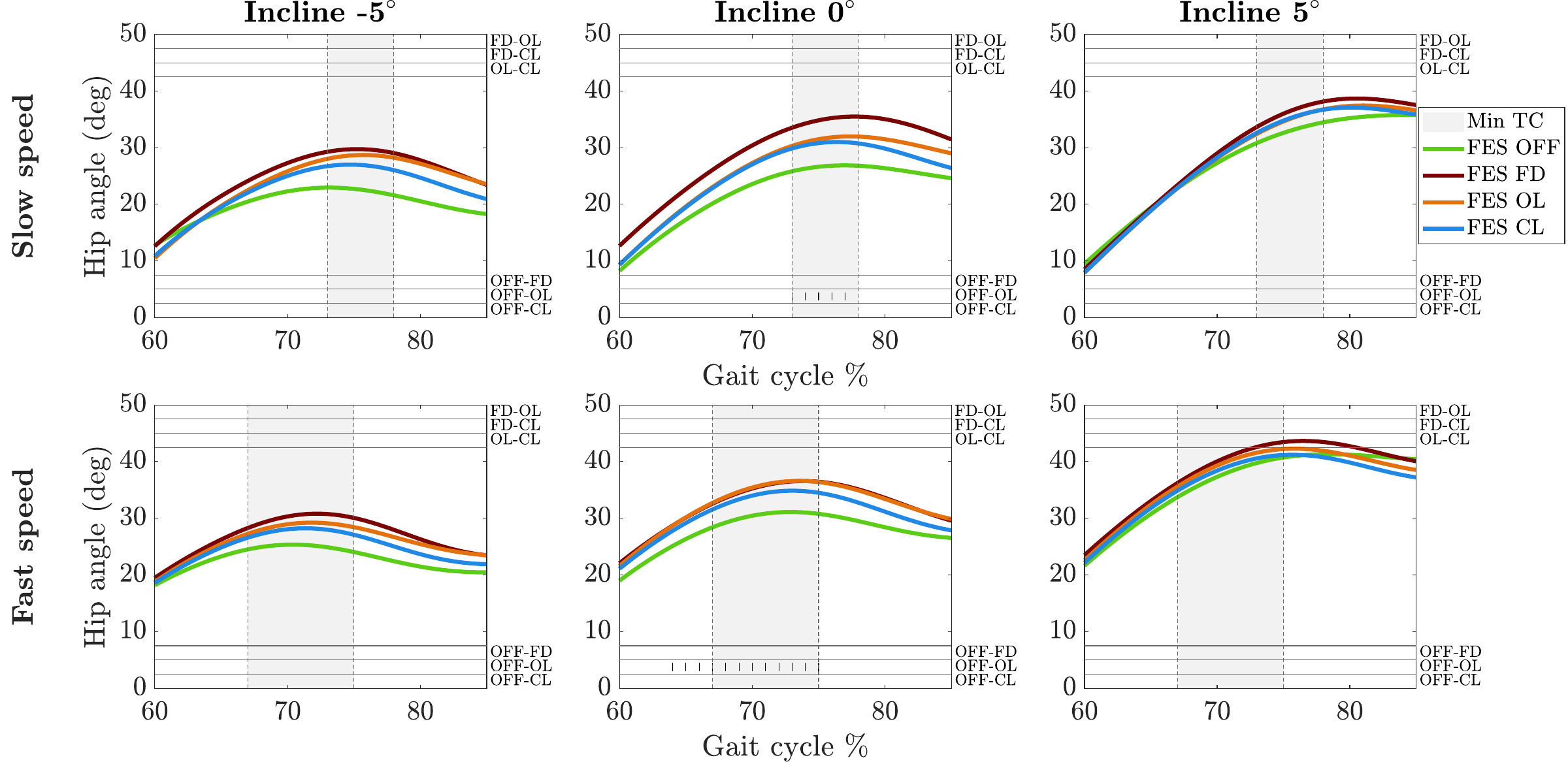}    
    \caption{Mean hip flexion during the swing phase for all subjects under the different conditions including toe clearance during slow walk (top row) and fast walk (bottom row), toe clearance during decline (left column), level walking (middle column) and incline (right column), and toe clearance during FES OFF (in green), FES FD (in red), FES OL (in orange) and FES CL conditions (in blue). The region where minimum toe clearance (Min TC) is expected, is highlighted in a grey background. Hip angle variability is not shown for clarity. Vertical bars ($|$) indicate points in the gait cycle where mean differences between conditions are statistically significant (p$<$0.05).}
    \label{fig:Hip Angle}
\end{figure*}

Figure \ref{fig:ToeClearance} shows the participants' recorded toe clearance\footnote{Due to the calculation of toe clearance using a 3D scan of a shoe with a flexible outsole, toe clearance during the stance phase appears to be negative.}. Statistically significant differences in toe clearance were observed during fast walking in the swing phase between the baseline (FES OFF) and the conditions FES OL and FES CL. An increase in toe clearance is observed mostly during the initial swing phase. Moreover, it can be seen that the mean toe clearance of the conditions FES OL and FES CL, is in almost all cases higher than the mean toe clearance observed during FES FD for the most part of the swing phase. This difference was statistically significant in some cases between FES FD and FES OL during the terminal swing. This may suggest that participants were able to adjust their gait in response to FES, or in anticipation of FES-induced perturbations \cite{Gordon2023b}, to ensure adequate toe clearance. This highlights the robustness of toe clearance as a kinematic metric during gait in healthy individuals (Table \ref{tab:toe_clearance_variability}).

\begin{table}[ht]
\centering
\caption{Mean within-cycle variability (mean SD$\pm$SD of the SDs) of toe clearance (all cycles) during swing and under different FES conditions, gait speeds, and inclinations. Units: mm.}
\label{tab:toe_clearance_variability}
\begin{tabular}{llccc}
\toprule
\textbf{Gait Speed} & \textbf{Condition} & \textbf{Incline -5°} & \textbf{Incline 0°} & \textbf{Incline +5°} \\
\midrule
\multirow{4}{*}{0.7 m/s} 
& FES OFF & 11.8 ± 2.7 & 11.5 ± 2.8 & 9.2 ± 2.7 \\
& FES FD & 11.0 ± 3.7 & 9.3 ± 2.4 & 10.1 ± 2.7 \\
& FES OL & 12.5 ± 2.7 & 12.2 ± 3.1 & 12.0 ± 2.6 \\
& FES CL & 11.8 ± 2.4 & 11.6 ± 2.6 & 11.6 ± 2.9 \\
\midrule
\multirow{4}{*}{1.2 m/s} 
& FES OFF & 15.6 ± 2.8 & 15.2 ± 2.4 & 11.8 ± 2.7 \\
& FES FD & 16.1 ± 4.0 & 14.4 ± 2.4 & 13.3 ± 2.9 \\
& FES OL & 16.8 ± 3.0 & 15.8 ± 2.6 & 15.0 ± 3.0 \\
& FES CL & 16.7 ± 3.1 & 15.6 ± 2.2 & 15.0 ± 2.9 \\
\bottomrule
\end{tabular}
\end{table}
Figure \ref{fig:Ankle Angle} shows the recorded ankle angle (positive is dorsiflexion). Due to high variability in the kinematics of the ankle joint, no statistically significant differences were found in the ankle dorsiflexion during the swing phase among the different conditions. Yet, when comparing the baseline to the FES FD condition, an increased plantarflexion is observed, which may be due to the stimulation of the GS muscles. From Figure \ref{fig:Ankle Angle} it can also be observed that the average ankle dorsiflexion is higher when the TA is stimulated (FES OL and FES CL) compared to when only the GS muscles are stimulated (FES FD). These observations verify that the intensities used to stimulate the calf muscles were sufficiently high to induce foot drop, and that the intensities used to stimulate the TA were sufficiently high to reduce foot drop. 

The effect of stimulation in the kinematics of the knees and the hips in the sagittal plane can be seen in Figures \ref{fig:Knee Angle} and \ref{fig:Hip Angle}, respectively. In Figure \ref{fig:Knee Angle}, it can be seen that during the FES FD condition, the participants had a slightly increased knee flexion compared to the baseline. During swing, this increased knee flexion is expected to increase toe clearance. A similar pattern was observed for the cases where FES was also applied on the TA. This deviation from the baseline was statistically significant for parts of the cycle only when the open-loop controller was used, suggesting that the increased toe clearance observed during FES OL may be partly attributed to increased knee flexion. From Figure \ref{fig:Hip Angle} we can observe a similar pattern for hip motion. When comparing against the baseline, hip flexion appears to be larger during the FES FD, FES OL and FES CL conditions. Significant differences in hip flexion are noticed for parts of the swing phase between the FES OFF condition and the FES OL condition, particularly during level walking. Significant differences in the hip and knee trajectories of the participants between the FES OL and the FES CL conditions were not observed. Yet, both the average hip flexion and the average knee flexion observed when the closed-loop controller was used appear to be smaller and closer to the baseline than the average hip and knee flexion observed during FES FD and FES OL. These changes suggest that our proposed closed-loop controller may lead to less compensatory mechanisms to guarantee adequate toe clearance, which can potentially lead to a healthier and more energetically efficient gait. 

\begin{figure*}
    \centering
    \includegraphics[width=0.95\linewidth]{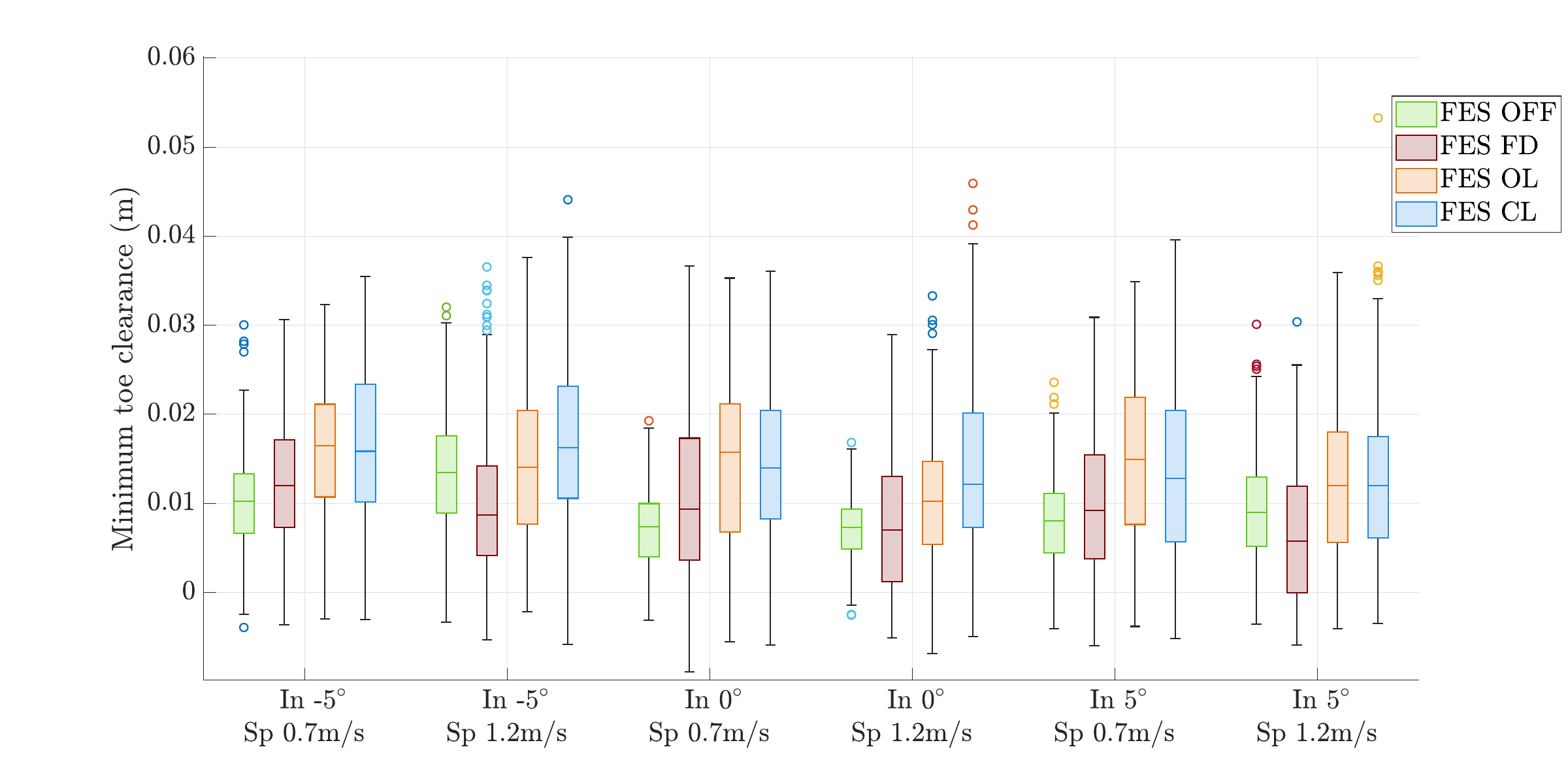}    
    \caption{Minimum toe clearance (MTC) for all subjects under the different conditions including MTC during slow walk and fast walk, decline, level walking, and incline, and during FES OFF (shown in green), FES FD (shown in red), FES OL (shown in orange) and FES CL conditions (shown in blue).}
    \label{fig:MinToeClearance}
\end{figure*}

In Figure \ref{fig:MinToeClearance} the effects of electrical stimulation on the minimum toe clearance (MTC) are summarised. The MTC is defined as the distance between the toes and the ground during the mid-swing phase, when the toes are closest to the ground. In the management of neurodegeneration, the MTC can be a reliable metric to quantify gait stability and safety. Looking at Figure \ref{fig:MinToeClearance}, it can be seen that even in the absence of a disturbance from FES, there have been recordings where the MTC was less than zero, indicating contact with the ground. Beyond musculoskeletal conditions which can cause this, foot contact with the ground in healthy individuals is not well documented and could potentially be a result of various reasons, including but not limited to fatigue, psychological factors, habit, or natural gait variability. It can also be noticed that while some differences in the average MTC were recorded between the different conditions, there were no statistically significant differences between the different FES modalities within each walking scenario. Yet, a slightly elevated mean can be noticed in the conditions where FES was applied on the TA (FES OL and FES CL) compared to the FES FD condition. This increased MTC observed when stimulating the TA appears to be consistent when comparing the results obtained from walking at different speeds and inclinations. When comparing the effect of the two corrective FES controllers on the MTC, the two controllers appear to be equally effective at increasing the MTC.   
\begin{figure*}
    \centering
    \includegraphics[width=0.95\linewidth]{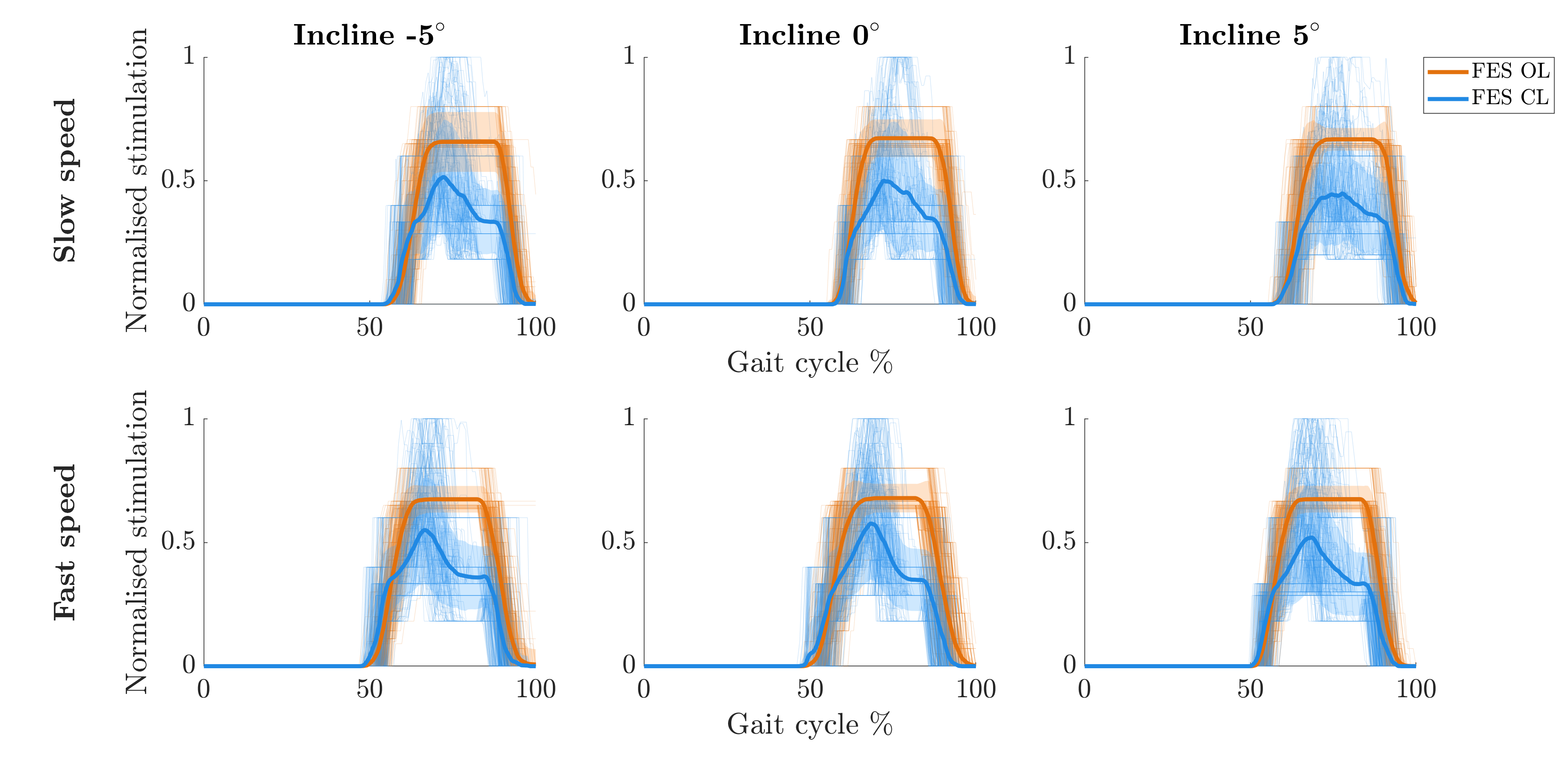}    
    \caption{Mean normalised stimulation intensity ($\pm$SD) for the anterior tibialis when using the open-loop FES controller (shown in orange) and the closed-loop FES controller (shown in blue) under the different conditions including slow walk (top row) and fast walk (bottom row), as well as decline (left column), level walking (middle column) and incline (right column). The individual stimulation profiles for each cycle are shown in the background.}
    \label{fig:Stimulation profiles}
\end{figure*}

\begin{figure}
    \centering
    \includegraphics[width=\linewidth]{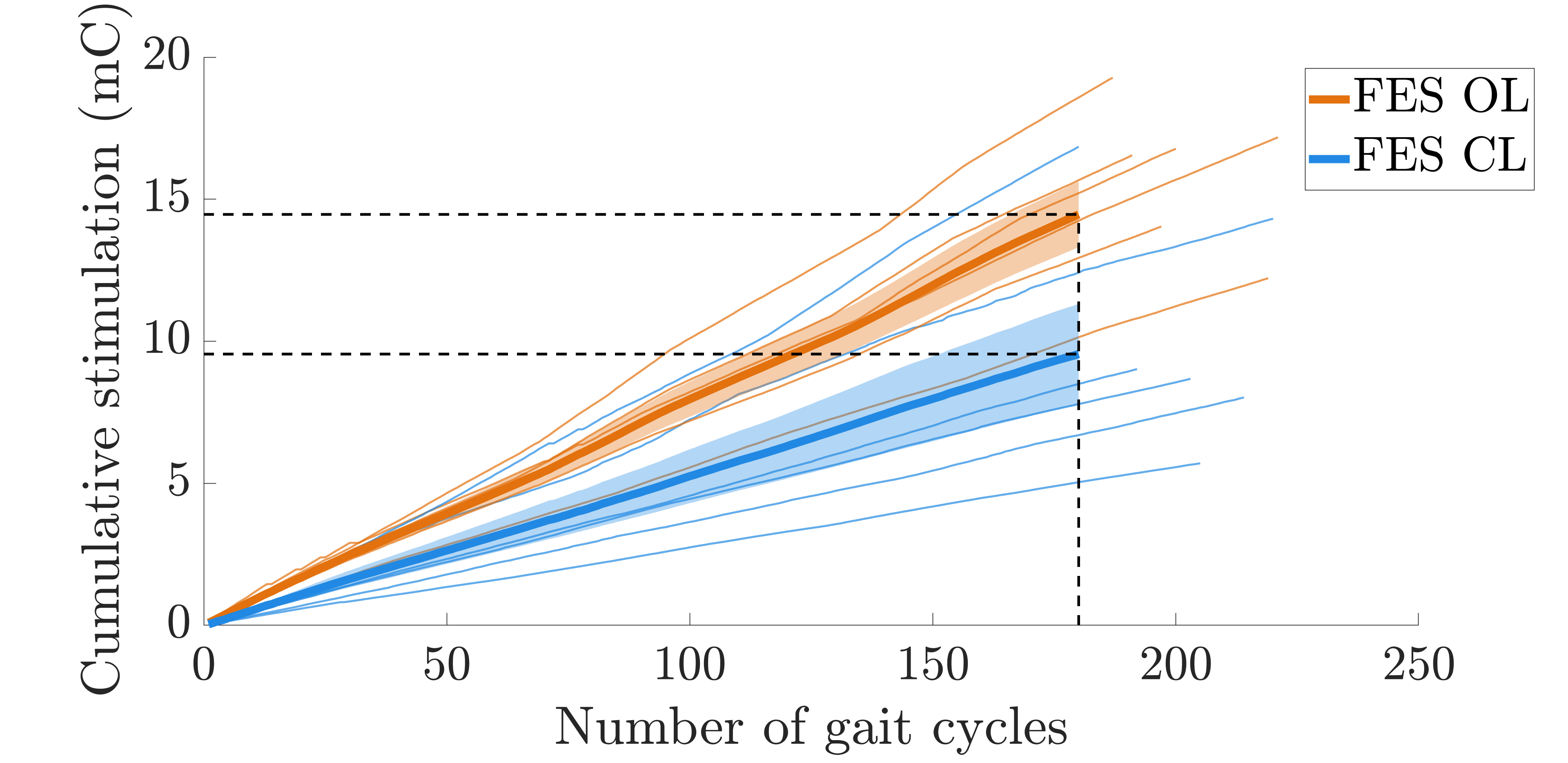}
    \caption{Cumulative sum of stimulation, expressed as total electrical charge (in mC), received by subjects for the open-loop FES controller (shown in orange) and the closed-loop FES controller (shown in blue).}
    \label{fig:Cumulative Stimulation}
\end{figure}

Lastly, the details regarding the amount of stimulation delivered by each corrective FES controller are shown in Figures \ref{fig:Stimulation profiles} and \ref{fig:Cumulative Stimulation}. Figure \ref{fig:Stimulation profiles} shows the average stimulation profile provided by each controller, normalised by the maximum stimulation intensity identified for each subject. It can be seen that the proposed closed-loop controller led to overall lower mean intensities than the open-loop controller. Even though the closed-loop controller needed, at times, to increase the stimulation intensity enough to reach the maximum stimulation intensity, the average profile generated by the closed-loop controller appears to be lower than the average profile generated by the open-loop controller for the whole duration of the swing phase. The stimulation profiles generated by the closed-loop controller appear to have a spike during the mid-swing phase where the MTC is expected and a more subtle spike during the terminal-swing phase where the heel strike is expected. This leads to muscle stimulation, which is reduced by an average of 34\% as it can be seen from Figure \ref{fig:Cumulative Stimulation}, which summarises the cumulative sum of stimulation, expressed as total electrical charge (in mC), received by the participants over the different conditions. 








\section{DISCUSSION}
The proposed closed-loop controller appears to significantly reduce the stimulation delivered to the TA muscles compared to the open-loop controller, without compromising toe clearance. This reduction was quantified at 34\% and is expected to decrease muscle fatigue and improve user acceptance. Notably, the controller's design includes a lower bound on stimulation during swing equal to $u^{TA_R}_{thr}$, which limits the maximum achievable reduction to 51\%. A less constrained design, could further reduce stimulation, but at the cost of reduced controller responsiveness. 

Interestingly, although the overall stimulation was lower, there were several gait cycles in which the closed-loop controller delivered more stimulation than the open-loop controller. This dynamic adjustment indicates that the closed-loop system may be more responsive to gait variations, which could arise from voluntary effort or changes in muscle force production due to stimulation.  As a result, the controller may also be better equipped to respond to FES-induced muscle fatigue, which is a persistent challenge in clinical use, while maintaining sufficient toe clearance to ensure safety.

A limitation of the proposed FES controller is the complexity involved with the calculation of toe clearance. Accurately estimating toe clearance may complicate the implementation of the proposed controller in the real world. However, it has been shown, that with the use of inertial sensors, it is possible to train data-driven models to obtain accurate estimates of foot clearance \cite{Fehr2024}. Thus, our next steps will involve this integration of the proposed FES controller with data-driven models that can accurately estimate toe clearance for the implementation of closed-loop FES in the real world and its evaluation on patients with foot drop. 

\section{CONCLUSIONS}
In this study, we introduced a novel closed-loop FES controller for managing foot drop, which dynamically adjusts stimulation intensity based on real-time toe clearance measurements. Unlike traditional open-loop systems with fixed or manually tuned intensities, our approach delivers stimulation only as needed, aiming to reduce the risks of overstimulation and understimulation. Using an experimental protocol involving healthy participants with induced foot drop, we evaluated  the performance of our proposed controller across varying walking speeds and inclines, and compared it against the performance of a conventional open-loop controller. The results show that while both controllers are equally effective in increasing toe clearance, our closed-loop controller significantly reduces the amount of stimulation delivered. This reduction has the potential to minimise muscle fatigue and improve user comfort, supporting the long-term usability and acceptability of FES in daily life.

\addtolength{\textheight}{-2cm}   




\section*{ACKNOWLEDGMENT}

The authors would like to thank all the people who took part in the experiment and the members of the SLMC group.
\bibliographystyle{IEEEtran}


\end{document}